\documentclass[letterpaper]{article}

\usepackage{natbib,alifeconf}  %% The order is important
\usepackage[hyphens]{url}

\title{Past Visions of Artificial Futures\\
\vspace{1mm}\Large{\emph{One Hundred and Fifty Years under the Spectre of Evolving Machines}}}
\author{Tim Taylor$^{1,2}$ \and Alan Dorin$^{1}$ \\
\mbox{}\\
$^1$Faculty of Information Technology, Monash University, Australia \\
$^2$Independent Researcher, Edinburgh, Scotland \\
tim@tim-taylor.com} % email of corresponding author

\begin{document}
\maketitle

\begin{abstract}
The influence of Artificial Intelligence (AI) and Artificial Life
(ALife) technologies upon society, and their potential to
fundamentally shape the future evolution of humankind, are topics very
much at the forefront of current scientific, governmental and public
debate. While these might seem like very modern concerns, they have
a long history that is often disregarded in contemporary
discourse. Insofar as current debates do acknowledge the history of
these ideas, they rarely look back further than the origin of the
modern digital computer age in the 1940s--50s.
In this paper we explore the earlier history of these concepts.
We focus in particular on the idea of self-reproducing and evolving
machines, and potential implications for our own species. We show that
discussion of these topics arose in the 1860s, within a decade of the
publication of Darwin's \emph{The Origin of Species}, and attracted
increasing interest from scientists, novelists and the general
public in the early 1900s. After introducing the relevant work from
this period, we categorise the various visions presented by these
authors of the future implications of evolving machines for
humanity. We suggest that current debates on the co-evolution of
society and technology can be enriched by a proper appreciation of the
long history of the ideas involved. 
\end{abstract}

\section{Introduction}

\begin{quotation}
``And why should one say that the machine does not live? It breathes
\ldots\ It moves \ldots\ And has it not a voice? \ldots\ 
And yet the mystery of mysteries is to view
machines making machines; a spectacle that fills the mind with
curious, and even awful, speculation.'' 

\hfill \emph{Coningsby} \cite[p.\ 154]{Disraeli:Coningsby}
\end{quotation}

By the climax of the British Industrial Revolution in the
early 1800s, the widespread introduction of increasingly sophisticated
manufacturing machines had raised anxiety about the
potential long-term consequences of mechanisation. Areas of unease
included not just the impact of technology on the labour conditions of
working people---a driving concern of the Luddite movement
\citep{Archer:Social}, but also the growing appreciation of the
\emph{self-amplifying} potential of the new machines. 
In 1844, the British author and future prime minister Benjamin
Disraeli wrote the novel \emph{Coningsby}. In a section describing the
industrial landscape of Manchester, the narrator raises the idea 
of \emph{machines making machines} and alludes to the profound
potential of such a development (see quote above).

During the same period, the scientific understanding of the complexity
of biological life was undergoing a revolution, in the
theories being developed by Charles Darwin and Alfred Russell
Wallace. Both theories were first presented at the Linnean Society of
London in 1858 \citep{Darwin:Tendency}, with a greatly extended
presentation of Darwin's theory appearing a year later with the
publication of \emph{The Origin of Species} \citep{Darwin:Origin}. 

At this time, the intellectual elite of England were a
richly connected web of thinkers, among whom ideas of science,
philosophy, technology, literature and the arts freely flowed. It did
not take long for the contemporaneous ideas of machines making
machines, and of the evolution of biological organisms, to be
connected---the result was the development of the idea of
\emph{self-reproducing and evolving machines}.

In this paper we explore the work of prominent authors of
the nineteenth and early twentieth centuries who addressed this
topic.\footnote{The history of the idea of self-reproducing machines
  dates back even earlier \citep{Taylor:Second}, but here we focus on
  machines that can both self-reproduce and evolve. We acknowledge
  that our literature search has been conducted primarily in English,
  and there may be relevant sources in other languages that we are
  unaware of. The review section of this paper draws upon material presented in
our new book \citep{Taylor:Second}.} We then identify
common themes in their work in terms of the implications of these
ideas for the future of human society and evolution, and conclude with
brief comments about the relevance of this work to current debates.

\section{Early writing on self-reproducing and evolving machines}

\subsection{Late Nineteenth Century (1860s--1890s)}

Almost as soon as \emph{The Origin of Species} was published, some
authors began exploring the applicability of Darwin's ideas to human
technology, and the potential consequences that this might entail.
% The most prominent early example is found in the work of
% Samuel Butler, with Alfred Marshall providing another instance of such
% thinking in the 1860s.

\subsubsection{Samuel Butler: \emph{Darwin Among The Machines} (1863)
  and later works}
As a young man, the English author Samuel Butler (1835--1902) spent
five years working in New Zealand.
%\citep{Taylor:Second}.
Shortly after his arrival in 1859 he read---and was greatly influenced
by---the recently published \emph{Origin of Species}. During his stay
he published a number of letters relating to Darwin's theory 
in the local Christchurch newspaper, \emph{The Press}. The second of
these,
%appearing
which appeared
in the 13 June 1863 edition under 
the pseudonym \emph{Cellarius}, was entitled 
\emph{Darwin Among the Machines} \citep{Butler:DarwinATM}.

Butler began the letter by noting the rapid pace of development of
machinery from the earliest mechanisms to the most sophisticated
examples of the day.
He commented that this had far outstripped the pace of development in
the animal and vegetable kingdoms, and asked what might be the
ultimate outcome of this trend. Observing the increasingly
sophisticated ``self-regulating, self-acting power'' with which
machines were being conferred, Butler suggested that humans ``are
ourselves creating our own successors.'' He further speculated that,
freed from the constraints of feelings and emotion, machines will
ultimately become ``the acme of all that the best and wisest man can
ever dare to aim at,'' at which point ``man will have become to the
machine what the horse and the dog are to man''
\citep{Butler:DarwinATM}. 

At that stage, Butler reasoned, the machines would still be
reliant upon humans for feeding them, repairing them, and producing
their offspring, and hence they would likely treat us kindly.
``[Man] will continue to exist, nay even to improve, and
will be probably better off in his state of domestication under the
beneficent rule of the machines than he is in his present wild
state.'' However, he then introduced the possibility of a time when
``the reproductive organs of the machines have been developed in a
manner which we are hardly yet able to conceive,'' noting that ``it is
true that machinery is even at this present time employed in begetting
machinery, in becoming the parent of machines often after its own
kind'' \citep{Butler:DarwinATM}.

Throughout his subsequent career, Butler wrestled with his views on
the application of Darwin's theory to machines, and the implications
for humanity.
In a subsequent letter to \emph{The Press} entitled \emph{Lucubratio
  Ebria}  \citep{Butler:Lucubratio}, published on 29 July 1865, he
presented a vision whereby machines are seen not as a competing
species, but rather as extensions to the human body. From this
perspective, Butler emphasised the capacity of machines to exert
positive evolutionary influences on the evolution of humankind, not
only by increasing our physical and mental capabilities, but also by
changing the environment in which we develop as individuals and evolve
as a species.

Upon his return to England in 1864, Butler continued to explore these
ideas. They appear in their most developed form in  
\emph{The Book of the Machines}, which constituted chapters 23--25 of
his novel \emph{Erewhon} \citep{Butler:Erewhon}.
Here he explored the collective reproduction of heterogeneous
groups of machines, rather than the reproduction of individuals.
Butler likened a complicated machine to ``a city or society'' \citep[p.\
212]{Butler:Erewhon}, and asked ``how few of the machines are there
which have not been produced systematically by other machines?''
\citep[p.\ 210]{Butler:Erewhon}.
He invoked a number of biological
analogies, such as bee pollination and specialisation of reproductive
function in ant colonies, to argue that collective machine
reproduction is no less like-like than
the self-reproduction of individual machines.

In \emph{Erewhon} Butler further explored the idea, first addressed in
\emph{Lucubratio Ebria}, that humans and machines are
\emph{co}-evolving, in a process driven  by market economics. However,
in contrast to his earlier writing, he now feared that this might be
detrimental to humankind, with machines evolving by acting
parasitically upon their designers: ``[the machines] have preyed upon
man's grovelling preference for his material over his spiritual
interests'' \cite[p.\ 207]{Butler:Erewhon}. Humans, he argued, are
economically invested in producing machines with ever more
``intelligibly organised'' mechanical reproductive systems \citep[p.\
212]{Butler:Erewhon}:  

\begin{quotation}
``For man at present believes that his interest lies in that
direction; he spends an incalculable amount of labour and time and
thought in making machines breed better and better
\ldots\ and there seem no limits to the results of accumulated
improvements if they are allowed to descend with modification from
generation to generation.'' 

\hfill \emph{Erewhon} \citep[p.\ 212]{Butler:Erewhon}
\end{quotation}

As machines evolved to become ever more complex, Butler was 
concerned that they might ``so equalise men's powers'' that
evolutionary selection pressure on human physical capabilities would
be reduced to a level that precipitated ``a degeneracy of the human
race, and indeed that the whole body might become purely rudimentary''
\citep[p.\ 224]{Butler:Erewhon}.  
This concern about the consequences for the human race of entering
a long-term co-evolutionary relationship with machines is taken up by
a number of later authors, most notably J.\ D.\ Bernal, whose work
we discuss later.

\subsubsection{Alfred Marshall: \emph{Ye Machine} (c.\ 1867)}

%Samuel Butler was not the only person dreaming of self-reproducing and
%evolving robots in the 1860s.
%In 1867, the young Alfred Marshall
Contemporaneous with Butler, in 1867 the
%In 1867---during the same period as Butler's works---the
%During the same period as Butler's works, in 1867 the
young Alfred Marshall
(1842--1924) wrote a series of four papers that formed the basis of
%
%In 1867---at around the same time as Butler's works---the young Alfred Marshall
%(1842--1924) wrote a series of four papers that formed the basis of
talks at ``The Grote Club''---an intellectual debating society at the
University of Cambridge. 
%talks at ``The Grote Club'' (an intellectual debating society at the
%University of Cambridge).
His theme was the extent to which the
activities of the human mind could be accounted for in
%purely
physical
terms.
In the third paper, \emph{Ye Machine}, Marshall proposed a model
for the objective study of mechanisms capable of learning and
intelligent action \citep{Raffaelli:EarlyP2}. Inspired by
recent
scientific work
%developments
in psychology, he described
%the design of
a
mechanical device (a robot in today's terms) equipped with sensors,
effectors and
%inner
circuitry that would allow it to develop
progressively more sophisticated ideas and reasoning about its
interactions with the world. 

The brain of Marshall's robot consisted of ``an indefinite number of
wheels of various sizes'' connected by bands which would be
automatically tightened whenever two wheels moved at the same time
\citep[p.\ 116]{Raffaelli:EarlyP2}. The design therefore 
implements what would now be classified as a kind of associative learning.
He goes on to describe how such a machine might also learn through
receiving positive or negative feedback about its actions, and how it
might develop instincts to allow it to maintain desired states.
Although such instincts could arise from the robot's associative
learning mechanisms, Marshall also speculated:
\begin{quotation}
``Nay, further, the Machine \ldots\ might make others like
itself. We thus get hereditary and accumulated instinct. For these
descendants, as they may be called, may vary slightly, owing to
accidental circumstances, from the parent. Those which were most
suited to the environment would supply themselves most easily with
fuel, etc.\ and have the greatest chance of prolonged activity. The
principle of natural selection, which indeed involves only purely
mechanical agencies, would thus be in full operation.''

\hfill Alfred Marshall, \emph{Ye Machine}, c.\ 1867

\hfill \citep[p.\ 119]{Raffaelli:EarlyP2}
\end{quotation}

\emph{Ye Machine} and the other papers presented by Marshall at
The Grote Club in the late 1860s had a limited audience at the
time, and they were not published in the scientific literature until
1994 (courtesy of the efforts of the late Tiziano Raffaelli). However,
the ideas Marshall developed in these papers are clear antecedents of themes
in his later work.\footnote{Marshall changed focus in his subsequent
career, becoming one of the founding fathers of neoclassical
economics. 
In his influential book \emph{The Principles of Economics}
%, first
%published in 1890,
%Marshall
he
drew analogies between economics and
biology, arguing
%biology.
% He noted that both dealt with systems ``of which the inner
% nature and constitution, as well as the outer form, are constantly
% changing'' \citep[p.\ 772]{Marshall:Principles}, and, further, that the
% development of both biological and industrial organisations ``involves
% an increasing subdivision of functions between its separate parts on
% the one hand, and on the other a more intimate connection between
% them'' \citep[p.\ 241]{Marshall:Principles}. He argued
that ``[t]he
Mecca of the economist lies in economic biology''
\citep[p.\ xiv]{Marshall:Principles}.}

%\subsubsection{Mary Ann Evans:
%  \emph{Impressions of  Theophrastus Such} (1879)} The final
%work of Mary Ann Evans (published under her pen-name George Eliot)

%\subsubsection{George Eliot (Mary Ann Evans):
\subsubsection{George Eliot:
  \emph{Impressions of Theophrastus Such} (1879)}
% The final
% %work of Mary Ann Evans (published under her pen-name George Eliot)
% %published
% work of George Eliot (Mary Ann Evans)
% was a series of short essays by an imaginary scholar
% \citep{Eliot:Impressions}. In
% %Chapter 17,
% the chapter
% % entitled
In the following decade, George Eliot (Mary Ann Evans) published her
final work, a series of
short
essays
%written
by an imaginary scholar
\citep{Eliot:Impressions}.
%In the
% short
The
%short
chapter
%\emph{Shadows of The Coming Race}, Eliot
\emph{Shadows of The Coming Race}
is a dialogue covering
%covered
%addresses
%discusses
%briefly
%explored
themes
%issues
first raised by Butler
regarding
the possibility of
machines developing the capacity for
self-reproduction and evolution by natural selection.
%The chapter
%She
It
also
touches
%touched
upon the potential consequences for humans, including mass
unemployment and an evolutionary degeneration of the mind and body.
Asked where
these ideas had come from, the narrator
explains
%explained
that ``[t]hey seem to
be flying around in the air with other germs.'' By the
%end of the nineteenth century
late 1800s
these topics were indeed very much in the air.\footnote{Butler
  thought that Eliot had
%  plagiarized \emph{Erewhon}, although that is not a
%  universally held view \citep{Taylor:Second}.}
  ``cribbed'' \emph{Erewhon} in her work, but
  the reality is more complicated \citep{Taylor:Second}.}
%  that is too simplistic a view of her
%  work \citep{Taylor:Second}.}

\subsection{Early Twentieth Century (1900s--1950s)}

By the turn of the twentieth century, the pace of
technological development had created a more pressing need for
considering where such progress might ultimately lead us.
During this period, the exploration of potential
futures of humanity in a world shared with self-reproducing, evolving
machines was attracting a wider audience. Where Samuel Butler had led, 
other authors soon followed. Here we highlight some of the first
examples of novels and other literature exploring self-reproducing
machines from the early twentieth century, and also discuss
speculative scientific work from this period.

\subsubsection{E.\ M.\ Forster: \emph{The Machine Stops} (1909)}
\label{sct-forster}

E.\ M.\ Forster's short story \emph{The Machine Stops}
\citep{Forster:Machine} was his only work of science fiction. It is
now regarded as a classic of dystopian literature \citep[p.\
50]{Evans:Wesleyan}.  

The story depicts a future in which
humans live underground in personal accommodation where corporeal
needs are entirely satisfied by technology (the global, all-nurturing
``Machine''). This leaves them free to concentrate on intellectual
development, although it also renders them physically
degenerate. Forster describes the Machine's ``mending apparatus''
that fixes problems and performs self-repair functions, evoking an
early image of a machine with a self-maintaining
organisation. It is the collapse of this functionality, brought about
by the mending apparatus itself falling into disrepair, that brings
the story to an apocalyptic end. Forster refers in passing to the
Machine evolving new ``food-tubes'', ``medicine-tubes'',
``music-tubes'' and even ``nerve-centres'', but these ideas are not
further explored. 

Forster acknowledged the influence of Samuel Butler in his work
\citep{Forster:ABook}---the vision in \emph{The Machine Stops} of a
future where an increasing dependency upon machines leads to the
degeneracy of the human body certainly echoes some of Butler's concerns.
Forster's image of self-maintaining machines sustaining human life was
further developed 20 years later by J.\ D.\ Bernal (see below).

\subsubsection{Karel \v{C}apek: \emph{R.U.R.: Rossum's Universal Robots}
  (1920)}
\label{sct-rur}

Themes of machine (collective) self-reproduction are further developed
in Karel \v{C}apek's
%well known
play \emph{R.U.R.: Rossum's Universal
  Robots} \citep{Capek:RUR}. Published in 1920 and first performed in
1921, the play introduced the word ``robot'' into the English
language.
%In the play, the
The
robots were
constructed from biochemical components and designed to resemble
humans, but lacked ``superfluous'' capacities such as feelings or the capacity to
reproduce. They were mass-produced in a factory to replace
human workers with a cheaper, more productive alternative. Most of the
production at the factory was carried out by robots themselves, with
only the most 
senior positions filled by humans. However, the complex formula for
manufacturing the key ``living material'' was a closely-guarded
secret, recorded by the factory's founder (Rossum) before his death
and kept in a safe to prevent it from falling into the hands of
competitors or the robots themselves.

One of the scientists in the factory experiments in
making robots with more human-like feelings such as pain and
irritability, but this results in unintended and ultimately disastrous
consequences when the robots come to despise their human masters and
rise up against them. This eventually leads to a stand-off where the
robots surround the factory and the people within it. The humans
realise that their only bargaining chip is the 
document that explains Rossum's formula, without which the robots
would be unable to produce more of themselves and would therefore die
out as a race when the current models fail.

The climax of the play thus revolves around a struggle for the
ownership of the written instructions that would allow the robots to
collectively produce more of themselves---a struggle for the ownership of the
robot's DNA, as it were.
% \footnote{As it turns out in the play, this
% bargaining strategy of the humans comes to nothing when they
% realise that Rossum's document has already been destroyed. The play
% ends when the last remaining human, Alquist, meets two robots who
% seem to have developed the capacity for love and human-like
% reproduction, thereby giving hope that although the human race is
% about to die out, human-like life will continue.}
This idea of the collective reproduction of a society of robots
reflects some of Butler's earlier ideas in \emph{Erewhon}.

\subsubsection{Early American Science Fiction (1920s-1950s)}

The appearance of American pulp science fiction magazines in the
1920s, and their growing popularity over the decades that followed,
provided a medium in which many writers explored the idea of
self-reproducing robots and evolving machines.
Perhaps the first example in this genre was the British writer S.\
Fowler Wright's story \emph{Automata}, published in the American
magazine \emph{Weird Tales} \citep{Wright:Automata}. With
echoes of Samuel Butler, the story extrapolates
the observed accelerating pace of technological development of the
time into the far future, to a point when machines no longer rely on
humans to service them. The machines become not only self-reproducing,
but also able to \emph{design} their own offspring. The story views
the takeover by machines as the inevitable next stage of evolution,
and serves as a warning of the unpredictable long-term consequences of
machine evolution: 

\begin{quotation}
``Even in the early days of the Twentieth Century man had stood in
silent adoration around the machines that had self-produced a
newspaper or a needle \ldots\ And at that time they could no more have
conceived what was to follow than the first ape that drew the
sheltering branches together could foresee the dim magnificence of a
cathedral dome.''

\hfill \emph{Automata} \cite[p.\ 344]{Wright:Automata}
\end{quotation}

Three years later, in 1932, the influential American sci-fi writer
and editor
John W.\ Campbell published \emph{The Last Evolution}
\citep{Campbell:LastEvolution}, which also anticipated the eventual
replacement of the human race by self-reproducing and self-designing
machines. However, 
Campbell's story is more optimistic than Wright's, 
foreseeing a period where humans live in peaceful and co-operative
coexistence with intelligent machines, with human creativity
complementing machine logic and infallibility. The end of the human
race comes not at the hands of the intelligent machines, but when a
species from another solar system invades Earth. The invasion prompts
the machines to design a new super-intelligent machine to thwart the
attack, and this itself spawns further rounds of creation of more
sophisticated machines---the final instantiation of which succeeds in
repelling the invaders but is ultimately the only surviving species on
Earth. Earlier in the story, the last two surviving humans console
themselves while contemplating their fate:
\begin{quotation}
``I think \ldots\ that this is the end \ldots of man \ldots\ But not
the end of evolution. The children of men still live---the machines
will go on. Not of man's flesh, but of a better flesh, a flesh that
knows no sickness, and no decay, a flesh that spends no thousands of
years in advancing a step in its full evolution, but overnight leaps
ahead to new heights.''

\hfill \emph{The Last Evolution} \cite[p.\ 419]{Campbell:LastEvolution}  
\end{quotation}

Campbell's vision of a complementary coexistence of humans
and intelligent machines is replaced by a less positive image in his
1935 story \emph{The Machine} (written under the pseudonym Don A.\ Stuart)
\citep{Stuart:Machine}. In the story a human-like race on a distant
planet design a thinking machine that is set the task of making better
versions of itself. The outcome is a machine that takes care
of all of the race's basic needs.
However, this ultimately leads to the degeneration of the race's
intelligence, civility, and its ability to look after itself---a
similar fate to those described by Butler in \emph{Erewhon} and
Forster in \emph{The Machine Stops}.
The machine decides that its presence has become detrimental 
to the planet's inhabitants, for they are not engaging with it
appropriately, but instead treating it like a god. The machine
resolves to leave the planet so that they can learn to live
independently once more.  

Laurence Manning's \emph{The Call of the Mech-Men}
\citep{Manning:Call} also mirrors ideas first aired by Butler 60 years
earlier. Two explorers discover a group of
extraterrestrial robots who have been living in underground caverns on
Earth since their spaceship was damaged many tens of thousands of
years earlier. The robots are amused when they
hear of humankind's view of itself as master of its technology,
remarking (in their stilted English):
``Machine gets fed and tended under that belief! Human even builds new
machines and improves year by year. Machines evolving with humans
doing all work!'' \cite[p.\ 381]{Manning:Call}.

Recurring themes of machine evolution and self-reproduction are seen
in stories over the following years. An example is Joseph E.\
Kelleam's \emph{Rust}, set on a post-apocalyptic
Earth where human-designed robots have survived after humankind has
been wiped out \citep{Kelleam:Rust}. The robots try to design and build
more of their kind before they succumb to erosion, but ultimately fail
in their attempts.
In Robert Moore Williams' \emph{Robots Return}
\citep{Williams:Robots}, three robots from a faraway planet
travel to Earth in search of information about their origins
many thousands of years earlier. To their surprise,
they discover that they were originally designed by humans, and had
been sent into space to accompany their creators in escaping a dying
Earth.
The humans did not survive the mission, but
the robots did, settling upon a distant world; there, they reproduced
and ultimately evolved into their current state.
One further example is A.\ E.\ van Vogt's \emph{M~33 in Andromeda}, in
which a spaceship of human explorers overcome an extraterrestrial
intelligence the size of a galaxy by constructing a self-reproducing
torpedo-manufacturing machine \citep{VanVogt:M33}. 

The most explicit exploration of machine
self-reproduction and evolution in early science fiction is found in
Philip K.\ Dick's \emph{Second Variety}
\citep{Dick:Second}. The story is set on Earth at the end of a
long-running war between East and West, in which Western forces are
driven to design killer robots to turn the tide on the
battlefield. The robots are highly autonomous, with each generation
of design becoming more sophisticated, including powers 
of self-repair and self-manufacture.
%The robots
They
eventually become too
dangerous for the human designers to be anywhere near, and they are
left to reproduce by themselves. Similar to Wright's \emph{Automata}
and Campbell's \emph{The Last Evolution}, the robots in \emph{Second
  Variety} eventually develop the ability to design their own
offspring, and increasingly sophisticated and human-like species of
killer robots begin to emerge. Echoes of these earlier stories are
also seen when one of the human characters remarks ``It makes me
wonder if we're not seeing the beginning of a new species. \emph{The}
new species. Evolution. The race to come after man''
\citep{Dick:Second}.

Themes of machine self-repair, self-reproduction and evolution were
central to various subsequent works by Dick. Another notable example
is \emph{Autofac} \citep{Dick:Autofac}, which ends with a vision of
the seeds of self-reproducing manufacturing plants being launched into
space.

\subsubsection{J.\ D.\ Bernal: \emph{The World, The Flesh and the Devil} (1929)}
In addition to the fictional explorations of machine
self-reproduction and evolution described above, we also see continued
interest in these topics from scientists in the early 1900s.
John Desmond Bernal (1901--1971) was an influential researcher who
conducted pioneering work on structural crystallography. Later in his
career he also became interested in the origins of life
\citep{Bernal:Physical}. In addition 
to his experimental work, he wrote many works on science and
society; his first monograph, and yet perhaps his
most futuristic writing, was entitled ``\emph{The World, the Flesh and
  the Devil: An Enquiry into the Future of the Three Enemies of the
  Rational Soul}'' \citep{Bernal:World}.\footnote{Arthur C.\ Clarke
  later described it as ``the most brilliant attempt at scientific
  prediction ever made'' \citep[p.\ 410]{Clarke:Greetings}.}

In this work, Bernal discusses how one might examine the future of humanity
in a scientifically defensible way. After sign-posting the
methodological and intellectual dangers to be avoided,
and discussing the unavoidable limitations, he proceeds to
explore what might be said of the three major kinds of struggle facing
humanity: against the forces of nature and the laws of physics in
general (``the world''); against biological factors including ecology,
food, health and disease (``the flesh''); and against psychological
factors including desires and fears (``the devil'').

Writing before the advent of space travel, atomic energy or computers,
Bernal first tackles how humankind might overcome the challenges that
arise from the material world. He argues that limitations of land and
energy in the world will eventually compel us to colonise space: ``On
earth, even if we should use all the solar energy which we received,
we should still be wasting all but one two-billionths of the energy
that the sun gives out. Consequently, when we have learnt to live on
this solar energy and also to emancipate ourselves from the earth's
surface, the possibilities of the spread of humanity will be
multiplied accordingly'' \citep[p.\ 22]{Bernal:World}.
After discussing plausible technologies for powering a spaceship (both
to escape the earth's gravitational field and also when in outer space), he goes
on to imagine how humans might set up permanent space colonies.

Bernal proposes a ``spherical shell ten miles or so in diameter''
\citep[p.\ 23]{Bernal:World} which could provide a habitable environment
for twenty or thirty thousand inhabitants. After discussing how the
construction of a sphere might be bootstrapped from a basic design
built largely of materials mined from an asteroid,
Bernal continues with a description of the organisation of a mature
sphere. It is imagined as
``an enormously complicated single-celled plant''
\citep[p.\ 23]{Bernal:World} with a protective ``epidermis'', complete
with regenerative mechanisms to protect against meteorites, mechanisms
for the capture of meteoric matter to be used as raw material for the
growth and propulsion of the sphere, systems for energy
production from solar energy, stores for basic goods such
as solid oxygen, ice and hydro-carbons, and mechanisms for the production
and distribution of food and mechanical energy. The sphere would also
have mechanisms for recycling all waste matters, ``for it must be
remembered that the globe takes the place of the whole earth and not
of any part of it, and in the earth nothing can afford to be
permanently wasted'' \citep[p.\ 25]{Bernal:World}.

The inhabitants of these globes in space would not be isolated, but
would be in wireless communication with other globes and with the
earth. In addition, there would be a constant interchange of people
between the globes and the earth via interplanetary transport
vessels. Having set out how the globes might function to sustain life
as ``mini-earths'', Bernal imagines a yet more ambitious scenario:
\begin{quotation}
``However, the essential positive activity of the globe or colony
would be in the development, growth and reproduction of the globe. A
globe which was merely a satisfactory way of continuing life
indefinitely would barely be more than a reproduction of terrestrial
conditions in a more restricted sphere.''

\hfill \emph{The World, The Flesh and the Devil}

\hfill \citep[p.\ 27]{Bernal:World}
\end{quotation}
Hence, the globe is conceived of as a fully self-maintaining and
self-reproducing unit---what might now be described as an
\emph{autopoietic} organisation \citep{Maturana:Autopoiesis}. Bernal
discusses methods by which a globe might construct another globe, and
then envisages how an evolutionary pressure to explore might arise
among a population of globes:

\begin{quotation}
``As the globes multiplied they would undoubtedly develop very
differently according to their construction and to the tendencies of
their colonists, and at the same time they would compete increasingly
both for the sunlight which kept them alive and for the asteroidal and
meteoric matter which enabled them to grow. Sooner or later this
pressure \ldots\ would force some more adventurous colony to set out
beyond the bounds of the solar system.''

\hfill \emph{The World, The Flesh and the Devil}

\hfill \citep[p.\ 29]{Bernal:World}
\end{quotation}
The enormous challenges
%that would be faced in
of
travelling interstellar
distances are addressed, but Bernal argues that such a vision is
nevertheless reasonable to consider: ``once acclimatized to space
living, it is unlikely that man will stop until he has roamed over and
colonized most of the sidereal universe, or that even this will be the
end. Man will not ultimately be content to be parasitic on the stars
but will invade them and organize them for his own purposes''
\citep[p.\ 30]{Bernal:World}.

Moving next to the possibilities of how our own bodies might develop
in the distant future, Bernal imagines that we will increasingly
replace and augment body parts with synthetic alternatives. Turning
to the activities such advanced beings might pursue, Bernal suggests
that, among other important scientific questions,
there would surely be intensive further study of life processes, and
the creation of synthetic life. However, ``the mere making of life
would only be important if we intended to allow it to evolve of itself
anew \ldots\ [however] artificial life would undoubtedly be used as
ancillary to human activity and not allowed to evolve freely except
for experimental purposes'' \citep[p.\ 45]{Bernal:World}.

Bernal's vision of the relationship between the future evolution
of humans and machines is more symbiotic than the futures imagined
by Forster and \v{C}apek: ``Normal man is an evolutionary dead
end; mechanical man, apparently a break in organic evolution, is
actually more in the true tradition of a further evolution''
\citep[p.\ 42]{Bernal:World}.
This perspective is more in line with the ideas expressed by Butler in
\emph{Lucubratio Ebria}, and with those of sci-fi authors such as John
W.\ Campbell. Bernal sees the main barriers towards progress
in these areas arising from human psychology---in addition to having
the desire for progress, we must also ``overcome the quite real
distaste and hatred which mechanization has already brought into
being'' \citep[p.\ 55]{Bernal:World}.
Various ways
%in which such barriers may be overcome
of overcoming such barriers
are suggested, but
Bernal does not discount the alternative possibility that we
ultimately find ways of living a simpler yet more satisfying life that
is not occupied by science or art but more at one with
nature.\footnote{In contrast to Butler in \emph{Darwin Among The
    Machines}, who thought that mankind was already past the point of
  no return in technology to allow such a reversal.}
%He also considers a third possibility---``the most unexpected, but not
%necessarily the most improbable'' \citep[p.\ 56]{Bernal:World}---that
He also considers a third possibility, ``the most unexpected, but not
necessarily the most improbable'' \citep[p.\ 56]{Bernal:World}, that
%the
%future
%evolution 
%of humanity
human evolution
might diverge, with one race following the natural path
and another race following the intellectual and technological path.

\subsection{More Recent Work (1950s--present)}

The 1940s and, in particular, the 1950s saw the
emergence both of the first rigorous theoretical work on the design
of self-reproducing machines, and of the first implementations of
artificial self-reproducing systems in software and
in hardware \citep{Taylor:Second}. This has been accompanied by
continued public debate about the implications of the technology for
the long-term future of our species. The history of these ideas from
this period is more widely acknowledged in current discussions, so we
end our review of the early development of these ideas here.
Details of work in the 1950s and early 1960s, and pointers to more
recent developments, can be found in \citep{Taylor:Second}.

\section{Discussion}

As demonstrated in the preceding sections, the early history of
thought about self-reproducing and evolving machines unveils a diverse
array of hopes and fears.
These contributions demonstrate that current debates about the
implications of AI and ALife for the future development of
humankind are actually a continuation of a conversation that has been
in progress for at least a hundred and fifty years.
In this final section, we consider the main recurring themes
that have emerged in our review.

\paragraph{Takeover by intelligent machines}

The most prominent theme apparent in this work is the fear that
machines might evolve to a level where they
%eventually
displace humankind as the dominant intelligent species.
While some writers proposed more positive, co-operative alliances
between humans and machines (e.g. Butler, Marshall, Wright, Campbell,
Bernal), none was fully convinced by this outcome, and all discussed
less desirable possibilities elsewhere in their
work.\footnote{Marshall is a possible exception, although his goal was
  to propose a model of biological learning and intelligent behaviour
  rather than to predict the future of humankind.}

The idea that we ourselves are creating our own successors can be seen
in the work of Butler, Eliot, \v{C}apek, Wright and Campbell.
%Indeed, some
Some
saw this not as a development to be feared, but rather as
a way in which the reach of humankind might be extended beyond the
extinction of our species (e.g.\ \v{C}apek, Campbell [\emph{The Last
  Evolution}], and Williams).

Most saw the evolution of increasingly intelligent machines
as an inevitable process.
%Of the work reviewed here, only \v{C}apek
In the work reviewed, only \v{C}apek
engages significantly with the idea that humans might exert some
control over the robots' reproduction. Butler and Bernal thought this
could likely only be achieved by humans
%completely
forsaking the
development of technology altogether.

The idea of \emph{self-repairing} machines is present in the work of
Eliot, Forster, Campbell [\emph{The Machine}] and Bernal, and this is
indeed a theme in current evolutionary robotics research, e.g.\
\citep{Bongard:Resilient}, \citep{Cully:Robots}. In contrast, we are
unaware of any serious scientific investigation of the idea of
\emph{self-designing} machines, which appears in the work of Wright,
Campbell and Dick. These authors 
portray self-design as a route by which the pace of machine evolution
can accelerate---these works, and Butler's before, strongly foreshadow
current interest in the idea of the \emph{technological
  singularity}. The concept has attracted increasing interest and
speculation since the birth of the digital computer age, particularly
in recent years through authors such as \cite{Moravec:MindChildren},
\cite{Kurzweil:Singularity} and
\cite{Bostrom:Superintelligence}. However, such speculations date back 
at least two hundred years; for a good discussion of the history of
these ideas, see \citep{Eden:Singularity}.

\paragraph{Implications for human evolution}

Beyond the idea that machines might become the dominant intelligent
species, the reviewed works have explored a number of potential
implications of self-reproducing machines for the future direction of
human evolution.

In \emph{Erewhon} Butler envisaged that humans might become
weaker and physically degenerate due to reduced evolutionary selection
pressure brought about by all-caring machines. Eliot and Forster
foresaw a similar outcome.
In contrast, an alternative outcome explored by Butler [\emph{Lucubratio
Ebria}] and Bernal is that human abilities might become significantly
\emph{enhanced} by the incorporation of increasingly sophisticated
cyborg technology.

Several authors emphasised that humans and machines are engaged in a
\emph{co}-evolutionary process. In \emph{Lucubratio Ebria} Butler
suggests that this closely coupled evolution of humans and machines
might increase our physical and mental capabilities. In particular, he
suggests that intelligent machines change the environment in which
humans develop and evolve---foreshadowing the modern idea of
biological \emph{niche construction} \citep{OdlingSmee:Niche}.
In \emph{The Last Evolution} Campbell envisaged a positive outcome of
this co-evolution, with human creativity working in harmony with
machine logic and infallibility.
Butler in \emph{Erewhon}, however, was more dubious of the
process, conjuring an image of machines as parasites benefiting from
the unwitting assistance of humans in driving their evolution.

The significance of self-reproducing machines as a technology to allow
humankind to explore and colonise other planets is a theme covered in
various works. The properties of self-repair and multiplication by
self-reproduction are seen as essential for attempts to traverse the
immense distances of inter-stellar---or even
inter-galactic---missions. Bernal's vision is of self-repairing and
self-reproducing living environments to allow multiple generations of
humans to survive such journeys. Williams, and Dick [\emph{Autofac}],
have our robot successors making the journey in place of us.

\paragraph{Implications for human society}

In addition to imagining consequences for human evolution, these
authors also envisaged how human society and the lives of individuals 
might be affected by the existence of super-intelligent machines. 

The prospect of humans becoming mere servants to machines
%appears in
%several works, including
%Butler's \emph{Darwin Among The Machines} and the sci-fi of Wright and Manning.
%those of
was raised by
%the work of
Butler [\emph{Darwin Among The Machines}], Wright and Manning.
However, Butler suggests that this
might not necessarily be a detrimental development---the machines
would likely take good care of us, at least for as
long as they still rely upon humans for performing functions
relating to their maintenance and reproduction.
%Eliot was concerned
%that as machines evolved more intelligence, lesser-skilled humans would
%find themselves increasingly unemployed

Many of the works explore how humans might spend their time in a
world where all of their basic needs are taken care of by beneficent
machines. In Forster's work, humans engage in the exchange of ideas
and academic learning (mostly about the history of the world before
the Machine existed). Similarly, Bernal suggests that we would be free
to pursue science, but also other areas of uniquely human activity
including art and religion. Individuals in Campbell's \emph{The
  Machine} are chiefly occupied with playing physical games and
pursuing matters of the heart. They also develop an unhealthy
reverence to the Machine as a god, to the extent that the Machine
ultimately decides to leave that planet so that the humans can learn
to live independently again. 

Likewise, Butler [\emph{Erewhon}] and Bernal discuss the
possibility that humans might separate from machines at some point in
the future, although in their works, in contrast to Campbell's, this
is a decision made by the humans rather than the machines. Bernal also
considers the possibility that the human species might ultimately
diverge into two, with one group pursuing the path of technological
co-evolution, and the other rejecting technology and searching for a
simpler and more satisfying existence more at one with nature.

\paragraph{Conclusion}

Concern about the impact of self-reproducing and evolving machines on
human society and our future evolution has a surprisingly long
history. As we have shown, this dates back at least as early as the
Industrial Revolution in Britain, and gains momentum with the
publication of \emph{The Origin of Species}. Modern debates about the
implications of AI and ALife technology are the continuation of a
conversation that has been in progress for over 150 years.

%We should be open to the possibility of
% To be sure, there is a possible
There is a possible
dystopian bias in
%these works,
the works reviewed,
which
%are
were
predominantly written by young, white men
\citep{Roberts:FromHomer}. It is indeed true that the large-scale
mechanical self-reproducing machines envisaged by these
%early
authors
have not yet been realised. Nevertheless, technological
advances in recent years have made possible various alternative
manifestations of their ideas.
% Although the large-scale mechanical self-reproducing machines envisaged
% by these early authors have not yet been realised, technological
% advances in recent years have made possible various alternative
% manifestations of their ideas.
%implementations.
Computer viruses, nano-machines, manufactured
bacteria and other self-reproducing wetware:  all testify to the
%continued---and increasing---need
continued and increasing need
for careful thought in this area. 
The spectre of self-reproducing and evolving machines is still very
much with us.

%We hope we have inspired
%participants in contemporary debates to
%explore all that has already been said in this conversation.

%%%%%%%%%%%%%%%%%%%%%%%%%%%%%%%
% Retro-edits for SRM book:
%
% Butler paragraph "Upon his return to England..." has been rewritten
%  here - backport these changes to the book
%
%  Section on RUR has some rewrites
%
% Robots Return para - minor edits
% Autofac para - minor edits
%
% Also do backports of .bib edits - do git diff on current with
% original versions of files

% Quote from Robots Return to possibily include in book:

% ``... the future is built of material taken from the past, and how can 
% we build securely when we do not know what our past has been?''
% (RR p. 142)

% (Also re their evolution: ``their evolution during their eight
% thousand years had been rapid'' (RR p. 142))

%\section{Acknowledgements}
%
%\footnotesize
%The review section of this paper draws upon material presented in
%our new book \citep{Taylor:Second}.

\footnotesize
\bibliographystyle{apalike}
\bibliography{taylor-history} % replace by the name of your .bib file

\end{document}